\documentclass[times,twocolumn,final,authoryear]{elsarticle}
\usepackage{ycviu}

\usepackage{colortbl}
\usepackage{arydshln}
\usepackage{adjustbox}
\usepackage{textcomp}
\usepackage{placeins}
\usepackage{multirow}
\usepackage{amsmath,amssymb,amsfonts}
\usepackage{algorithmic}
\usepackage{algorithm}
\usepackage{graphicx}
\usepackage{subcaption}
\usepackage{textcomp}
\usepackage{placeins}
\usepackage{lipsum}
\usepackage{mathtools}
\usepackage{cuted}
\usepackage[table,dvipsnames]{xcolor}
\usepackage{times}
\usepackage{epsfig}
\usepackage{graphicx}
\usepackage{epstopdf}
\usepackage{enumitem}
\usepackage{fixltx2e}
\usepackage{array}
\usepackage{arydshln}
\usepackage{comment}
\usepackage[colorlinks=true]{hyperref}
\usepackage{fancyhdr}
\usepackage{textcomp}

\definecolor{newcolor}{rgb}{.8,.349,.1}

\journal{Computer Vision and Image Understanding}

\begin{document}

\clearpage

\thispagestyle{empty}

\ifpreprint
  \vspace*{-1pc}
\else
\fi






 




\clearpage
\ifpreprint
  \setcounter{page}{1}
\else
  \setcounter{page}{1}
\fi


\begin{frontmatter}

\title{A novel network training {approach} for open set image recognition}
\author[1]{Md Tahmid \snm{Hossain} \corref{cor1}}
\cortext[cor1]{Corresponding author [Paper under review]}
\ead{mt.hossain@federation.edu.au}

\address[1]{School of Science, Engineering and Information Technology, Federation University, Gippsland Campus, Churchill, VIC 3842, Australia}

\address[2]{Discipline of Information Technology, Murdoch University, Perth, WA 6150, Australia
}
\author[1]{Shyh Wei \snm{Teng}}

\author[1]{Guojun \snm{Lu}}
\author[2]{Ferdous \snm{Sohel}}

\received{1 May 2013}
\finalform{10 May 2013}
\accepted{13 May 2013}
\availableonline{15 May 2013}
\communicated{S. Sarkar}

\begin{abstract}
{
 Convolutional Neural Networks (CNNs) are commonly designed for closed set arrangements, where test instances only belong to some `Known Known' (KK) classes used in training. As such, they predict a class label for a test sample based on the distribution of the KK classes. However, when used under the Open Set Recognition (OSR) setup (where an input may belong to an `Unknown Unknown' or UU class), such a network will always classify a test instance as one of the KK classes even if it is from a UU class. {As a solution, recently, data augmentation based on Generative Adversarial Networks (GAN) has been used}. In this work, we propose a novel approach for mining a {`Known Unknown Trainer' or KUT} set and design a deep OSR Network (OSRNet) to harness this dataset. The goal is to teach OSRNet the essence of the {UUs} through KUT set, which is effectively a collection of mined ``hard Known Unknown negatives''. Once trained, OSRNet can detect the {UUs} while maintaining high classification accuracy on {KKs}. We evaluate OSRNet on six benchmark datasets and demonstrate it outperforms contemporary OSR methods.
 }
 \end{abstract}


\end{frontmatter}



\section{Introduction}
\label{sec:introduction}

\begin{figure}
\begin{center}
   \includegraphics[width=0.8\linewidth]{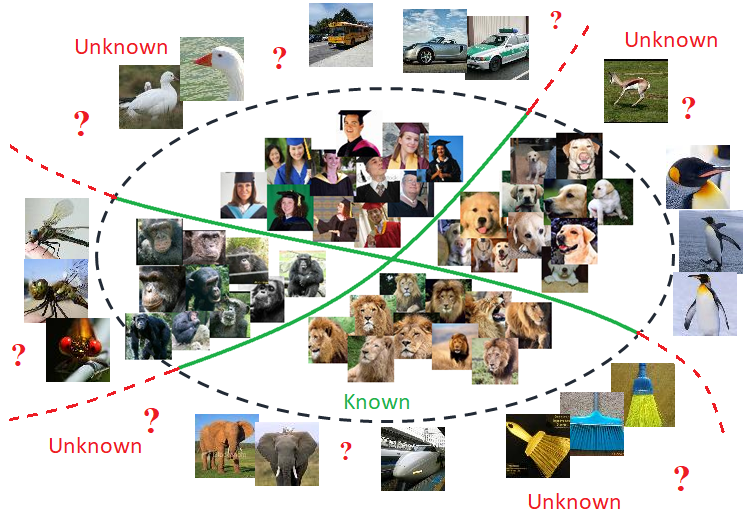}
\end{center}
   \caption{Conceptual illustration of a 4-class classifier decision boundary. A deep CNN can produce high-precision classification results so long a novel test instance belongs to one of the {KK} classes. However, when the CNN is exposed to {UU} instances, it misclassifies such images as one of the {KK} classes because of its lack of precision outside the {KK} distribution.}
\label{fig:simple}
\end{figure}

{An image classifier is expected to correctly classify images belonging to the `Known Known' (KK)} distribution seen during training. However, {during inference,`Unknown Unknown' (UU)}\footnote{Note that we refer to the collection of {Known Unknowns (KUs) and Unknown Unknowns ({UUs}) as unknowns.}} instances might trigger incorrect classification. As the training set comprises of a finite number of classes, identifying a {UU} instance is a challenge (see Figure \ref{fig:simple}). This is referred to as the Open Set Recognition (OSR) problem. To better understand the importance of OSR, let us consider an example. Imagine an autonomous car is trained to recognize 10 different street signs, and the car responds according to a pre-defined set of rules. The on-board sensors perceive the environment and feed data to the trained classifier (e.g., Convolutional Neural Network or CNN). In an environment full of objects beyond the closed set training classes, it is likely for a CNN to classify a non-street sign object as one of the 10 street signs (e.g., a commercial billboard might be incorrectly perceived as a stop sign). Such an event and the consequent pre-programmed response of the autonomous car can lead to an undesired situation, even to a fatal accident. 


{In traditional classification tasks, a deep network is usually trained only on a KK dataset \textit{$D\textsubscript{KK}$} without considering the unknowns. For an effective OSR, training a network with `everything else' in the world as the unknown is unrealistic. Therefore, a {Known Unknown Trainer (KUT)} dataset (referred to as {\textit{$D\textsubscript{KUT}$}} hereafter) is required for this task. {Generative Adversarial Network (GAN)-based artificial images have often been used as a data augmentation-based solution in a number of contemporary works \citep{perera2020generative,Neal1,Ge1}. However, being trained on only KK samples, the distribution of the augmented data is not always compatible with the {UU} distribution present in the test set.} Moreover, GANs suffer from a number of issues stemming from the generation techniques themselves, such as unwanted artefacts, and mode collapse \citep{goodfellow2016nips}. These deteriorate the network's performance and increase training time. Another way to address the OSR problem is to find an effective SoftMax threshold to reject {UU}s \citep{Bendale1,Oza1,Ge1}. }

 {In this work, we argue that drawing a boundary between the KK and the unknown is the key to efficiently handling the OSR task and we aim to approximate such a decision boundary by mining a {\textit{$D\textsubscript{KUT}$}}. To accomplish this, we propose a way of mining the hard {KU} negatives into {\textit{$D\textsubscript{KUT}$}} and design a deep network to be trained on this dataset. {\textit{$D\textsubscript{KUT}$}} is mined from publicly available benchmark datasets (\textit{$D\textsubscript{x}$}) (where, \textit{$D\textsubscript{KUT} \subset D\textsubscript{x}$}). 
Images from \textit{$D\textsubscript{x}$} inducing high probability ($P >$ some threshold $T$) for one of the {KK} classes gets admission into {\textit{$D\textsubscript{KUT}$}} (calculation of $T$ is explained in Section 3). 
We demonstrate that OSRNet, which is trained to distinguish {\textit{$D\textsubscript{KUT}$}} from \textit{$D\textsubscript{KK}$} can identify novel {UU} instances at test time. \textit{$D\textsubscript{KUT}$} does not include any of the classes present in the {UU} test fold to ensure a fair evaluation.}

 

 The proposed OSRNet has two parts: a traditional CNN as the base and a Confidence Subnetwork or CS. The CNN is trained conventionally to classify a given {\textit{$D\textsubscript{KK}$}}. The CS, which effectively is an Artificial Neural Network (ANN), is separately trained to identify {UU}s. Once CS is trained, it is augmented to the trained CNN, and OSRNet is formed (details in Section \ref{osnet}). 
  At inference time, the newly formed network works as one single end-to-end unit. Inside OSRNet, the CNN produces class predictions and at the same time, the CS outputs a single confidence score $S = [0,1]$ to indicate whether an instance belongs to one of the {KKs} or not ($S \xrightarrow{} 1$ denotes a high chance of the input being {UU}). A cut-off value $\delta$ is used on $S$ to determine the final outcome, i.e., whether to accept or reject the class label produced by the base CNN. OSRNet does not require any additional computation module (e.g., EVT) outside the deep network's perimeter. The entire inference process is end-to-end without any bottleneck.  

\begin{figure}
\begin{center}
   \includegraphics[width=0.95\linewidth]{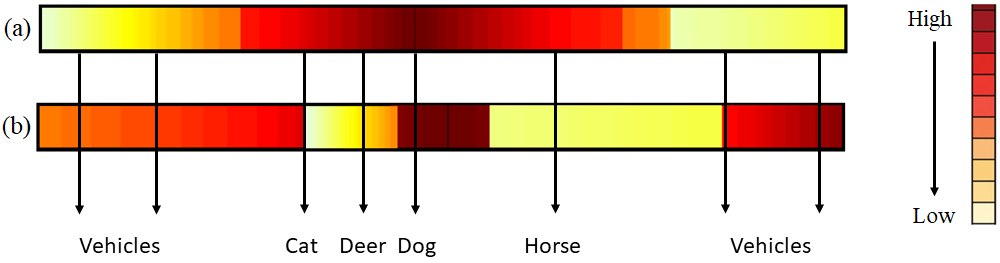}
\end{center}
   \caption{Behavioural difference of a trained CNN on classes it has and has not seen during training. For a {KK} test instance, usually, the correct class receives maximum probability, and the leftover energy is distributed to visually similar classes. In contrast, no such pattern is found for an Unknown test instance. \textbf{(a)} denotes the probability heat map for CIFAR-10 dog class (test set). Most of the energy is distributed to either dog or among classes visually similar to dog, e.g., cat, deer, and horse. In \textbf{(b)}, when unknown instances (taken from CIFAR-100) are misclassified as dog, the leftover energy distribution does not follow the same pattern.
   }
\label{fig:heatBar}
\end{figure}

{The} OSRNet architecture is inspired from the observation by Bendale et al. \citep{Bendale1} depicted in Figure \ref{fig:heatBar}. It is reported that when classifying an image from {\textit{$D\textsubscript{KK}$}}, traditional CNNs generally produce a high probability score for the correct class while the leftover probability is usually distributed across visually similar classes. Even when a CNN is unsure or misclassifies a {KK} instance, the probability scores mostly remain concentrated to visually similar classes. In contrast, when an unknown instance gets classified as one of the {KK} classes, the leftover probability distribution does not usually follow such a pattern \citep{Bendale1}. For illustration (in Figure \ref{fig:heatBar}), we train a simple CNN on CIFAR-10 and generate a probability heat map for the dog class test set. It is evident in Figure \ref{fig:heatBar}(a) that either dog or classes visually similar to dog receives most of the energy. In Figure \ref{fig:heatBar}(b), we feed the same CNN instances from unseen CIFAR-100 classes and produce the heat map with 1,000 samples misclassified as dog. The leftover probability distribution has marked contrast to Figure \ref{fig:heatBar}(a).
This is the pattern we aim to exploit in this work for OSR. Because of the above-mentioned behavioural difference on {KKs} and any Unknown instances, we expect the intermediate layer features (Fully Connected or FC) of a CNN (trained on {\textit{$D\textsubscript{KK}$}}) to reflect this difference. Therefore, FC features should be distinct for {KK} and unknown samples. Collecting such FC features for {\textit{$D\textsubscript{KK}$}} is straightforward from our base CNN. However, mining an appropriate {\textit{$D\textsubscript{KUT}$}} set to collect FC features is challenging. Our proposed method addresses both issues.

In this work, we make the following contributions:



 \begin{itemize}
 
 \item We propose a way of building the {Known Unknown Trainer} dataset {\textit{$D\textsubscript{KUT}$}} for OSR. 
 
\item To effectively distinguish between the {KKs} from the {UU}s at test time, a novel deep network (OSRNet) is proposed.

\item We extensively evaluate OSRNet and compare with contemporary OSR methods on six benchmark datasets. OSRNet not only detects {UU}s with higher precision but also exhibits impressive discriminative power within {\textit{$D\textsubscript{KK}$}}.

\item Finally, we discuss the underlying reasons behind the effectiveness of the proposed {\textit{$D\textsubscript{KUT}$}} and OSRNet in OSR.
\end{itemize}

The rest of the paper is organized as follows: Section \ref{related} discusses the relevant works. In Section \ref{osnet}, we demonstrate the architecture of the proposed OSRNet, {\textit{$D\textsubscript{KUT}$}} mining process, how an effective threshold $T$ is chosen as the selection criteria and why it works so well. In Section \ref{experiments}, we outline the datasets used for training and testing and implementation details. Section \ref{perform} provides an experimental evaluation of the proposed and existing methods. Section \ref{discussion} provides a discussion on the {Known Unknown Trainer} image set. Finally, Section \ref{conclusion} concludes this paper.

\section{Related Work}\label{related}
\textbf{Open Set Recognition. }
OSR methods can be categorized into two main types: CNN-based and non-CNN-based. Scheirer et al. \citep{open1,wsvm1} first formalized the OSR problem and proposed an SVM-based solution. Extreme Value Theory (EVT) is used in a number of works \citep{evt1,zhang2016sparse} to reject {UU} instances receiving probability score lower than a threshold. Dang et al. \citep{dang} proposed an OSR model where edge exemplars are selected for every class based on local geometrical and statistical properties \citep{li2}. Later, EVT-based rejection rule is adopted to reject any {UU} input that lies outside the {KK} class boundaries. Recently, deep learning-based networks are found to be more effective in OSR. 

OSR with Counterfactual Images (OSRCI) \citep{Neal1} uses GAN to produce Counterfactual Images (CI) by morphing instances from {\textit{$D\textsubscript{KK}$}} to an extent where they no longer are recognizable as a true class object (neither {KK} nor unknown). Later, these CI are used as the {\textit{$D\textsubscript{KUT}$}} (i.e., CI $\approx {\textit{$D\textsubscript{KUT}$}}$). An additional $(N+1)^\text{th}$ `other' class is introduced to accommodate images in {\textit{$D\textsubscript{KUT}$}} during training. At inference time, the classifier is expected to classify {UU} instances as `other'. Although the network is expected to classify real-world objects, CI used as {\textit{$D\textsubscript{KUT}$}} lack visual characteristics of natural images limiting the effectiveness of OSRCI.

Bendale et al. \citep{Bendale1} replaced the SoftMax function with OpenMax. It is argued that forcing a network's total output probability to sum up to 1 leads the network to put undue probability score to {UU} instances at test time. It has been reported that test instances from {\textit{$D\textsubscript{KK}$}} put high prediction scores on the true class while the leftover probability is distributed to visually similar classes. However, the output probability distribution does not exhibit the same pattern for unknown instances. Inspired from this observation, the penultimate layer features (Activation Vector or AV) are extracted, and a mean vector (AVM) is calculated for each class (class-wise images are fed to the CNN). At test time, the AV of an image is extracted, and its distances to all the AVMs are calculated. Later, these distances are used with a threshold to detect whether the image belongs to a {KK} class or not. 
Ge et al. \citep{Ge1} supplemented OpenMax with Generative OpenMax (G-OpenMax) where they used GAN-generated data for OSR training. 
Classification-Reconstruction learning for Open-Set Recognition (CROSR) \citep{Yoshihashi} is also an extension of OpenMax, but a different route is followed. A two-part deep network is used: a {KK} classifier and a {UU} detector. Multiple intermediate layers of the main CNN classifiers are treated as latent features, and a decoder is used to reconstruct the input. The {UU} detector and the classifier, both exploit the latent space features jointly to output detection decision and class label respectively.

{
Class Conditioned Auto-Encoder (C2AE) \citep{Oza1} also adopts an Encoder Decoder-based reconstructive approach. An encoder network is first trained on {\textit{$D\textsubscript{KK}$}}, and the penultimate layer is used as the encoded vector (EV). For each class in {\textit{$D\textsubscript{KK}$}}, one such EV is stored as the class condition vector. A decoder is later used to reconstruct the input from the EV. For each input to the encoder, the EV output is compared against all the stored EVs. The decoder reconstructs the input as perfectly as possible whenever a match is found between the output and stored EVs. However, for {UU} images (with no EV matches), the decoder is designed to perform a poor reconstruction so that the reconstruction error is high at inference time. C2AE is not an end-to-end unit and leaves room for further improvement.}

Geng et al. \citep{geng2020guided} proposed a visual and semantic prototypes-jointly guided CNN (VSG-CNN) to achieve the task of OSR. Instead of using traditional cross-entropy loss, a distance-based cross-entropy loss (DCE) is used to find out the probability of a test instance belonging to each {KK} class. Once this probability set is at hand, the overall entropy is calculated and based on a threshold, the test instance is either rejected as a {UU} class or classified as one of the {KK} ones.\\ 
\textbf{Out-of-Distribution Detection. } Out-of-Distribution (OOD) or Anomaly Detection is correlated to OSR. 
OSR deals with identifying real but {UU} images while classifying any {KK} instance. On the other hand, OOD detectors focus on anomalous outlier detection. Sometimes, these anomalies can be visually unrecognizable \citep{ngfooling1} or `rubbish'. Some methods only detect the outliers first, and a separate classifier is used later for classification only if an input is deemed as {KK} by the detector \citep{odin,vyas}.

\begin{figure}
\begin{center}
   \includegraphics[width=1\linewidth]{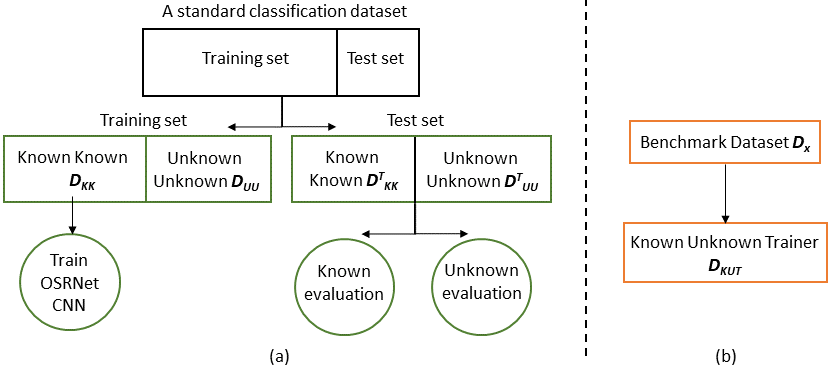}
\end{center}
   \caption{\textbf{(a)} An overview of the dataset splits in this work. \textbf{(b)} The {Known Unknown Trainer} dataset {\textit{$D\textsubscript{KUT}$}} is a subset of \textit{$D\textsubscript{x}$} mined following our proposed method. {\textit{$D\textsubscript{KUT}$}} does not contain any of the {\textit{$D\textsuperscript{T}\textsubscript{UU}$}}/{\textit{$D\textsubscript{UU}$}} classes. As a result, OSRNet remains blind regarding the classes it is going to encounter during evaluation. }
\label{fig:dsets}

\end{figure}

\begin{figure*}
\begin{center}
   \includegraphics[width = .86\linewidth]{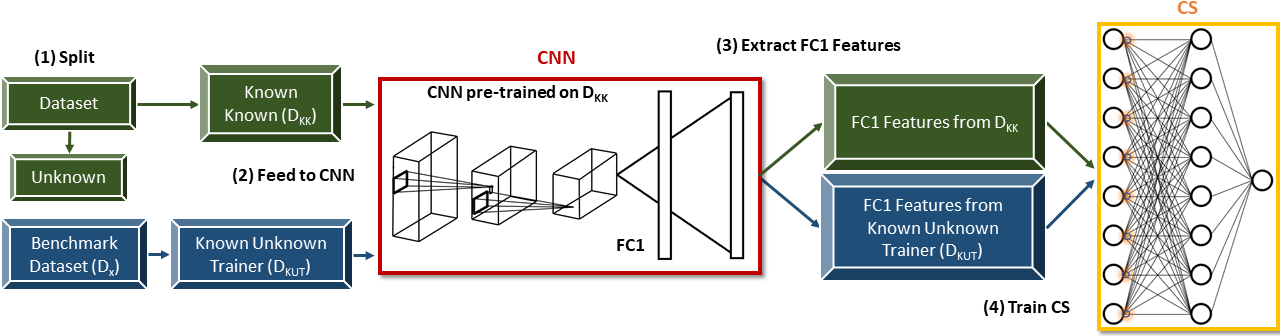}
\end{center}
   \caption{An overview of OSRNet training process. A CNN (\textit{ResNet{$\textsubscript{2FC}$}}) is trained on the {KK} split ({\textit{$D\textsubscript{KK}$}}) of a dataset. Images from {\textit{$D\textsubscript{KK}$}} and a {\textit{$D\textsubscript{KUT}$}} mined from another benchmark dataset \textit{$D\textsubscript{x}$} is fed to this already trained CNN. \textit{FC\textsubscript{1}} features are extracted for both {\textit{$D\textsubscript{KK}$}} and {\textit{$D\textsubscript{KUT}$}} and the CS is trained with these features and binary labels. Once trained, this CS is augmented to the corresponding CNN layer (\textit{FC\textsubscript{1}}) and the proposed OSRNet is formed. At inference time, OSRNet works as one single end-to-end unit. The augmented CS receives \textit{FC\textsubscript{1}} features from the CNN and detects whether the input belongs to one of the {\textit{$D\textsubscript{KK}$}} classes or not. The CNN, on the other hand, simultaneously classifies an instance without any interference from the CS.}
\label{fig:os2}
\end{figure*}

A common way to tackle the OOD problem is to simply augment an additional class to an existing CNN so that all OOD instances are classified as the `other' class \citep{Grosse1}. However, adding an additional class for all other images in the world do not perform consistently well on different benchmark datasets \citep{ngfooling1}. Hendrycks et al. \citep{HendrycksABaseline1} used the SoftMax probabilities as a heuristic to detect outlier images. A simple thresholding technique is applied based on the assumption that in distribution samples will always have higher probabilities, and OOD instances would not trigger high confidence predictions. However, this assumption is inaccurate. SoftMax thresholding does not work well as CNNs often misclassify an OOD image with high probability. As a solution, some detection methods introduce OOD samples in the training data and employ a custom loss function to uniformly diffuse probability on OOD training samples \citep{HendrycksDeepAnomaly1,hein1,Lee1,DeVries1}. These loss functions constrain the CNN from wrong overconfident predictions, but overall accuracy is compromised. {Moreover, These methods use one benchmark dataset as {\textit{$D\textsubscript{KK}$}} and other entire publicly available benchmark datasets \citep{HendrycksDeepAnomaly1} (\textit{$D\textsubscript{x}$}) as \textit{$D\textsubscript{KUT}$} (i.e., \textit{$D\textsubscript{x}$} $\approx$ {\textit{$D\textsubscript{KUT}$}}). We argue that mining only the hard known unknown negatives from \textit{$D\textsubscript{x}$} into {\textit{$D\textsubscript{KUT}$}} works better (i.e., {\textit{$D\textsubscript{KUT}$}} $\subset $\textit{$D\textsubscript{x}$}).}

Li et al. \citep{Li1} investigated the statistical properties of different CNN layer features to find a distinguishing pattern between In and Out of Distribution images. It is reported that the convolution outputs for {\textit{$D\textsubscript{KK}$}} and OOD instances have subtle difference. The difference is so subtle that even the most impactful dimensions in PCA (PCA head) fail to capture the difference. However, the tail (less informative eigen dimensions) of PCA shows a difference in the pattern. These features are used to train a cascade classifier for OOD detection.

GAN is used for OOD \citep{hein1,Mandal1,Lee1,Ge1,Neal1}. Lee et al. \citep{Lee1} used GAN to produce a {\textit{$D\textsubscript{KUT}$}} that neither belongs to {\textit{$D\textsubscript{KK}$}}, nor lies far away. A custom loss function (\textit{L}) based on Kullback-Leibler divergence between a uniform distribution \textit{U} and prediction on OOD instances is used for probability diffusion.


\section{Proposed Method}\label{osnet}
\subsection{OSRNet}
Before elaborating on the proposed method, we outline the dataset splits for convenience:
\begin{itemize}
    \item (\textbf{Training}) One {KK} set {\textit{$D\textsubscript{KK}$}}.
    \item (\textbf{Training}) One {Known Unknown Trainer}  set {\textit{$D\textsubscript{KUT}$}} mined from a \textit{$D\textsubscript{x}$} following our proposed method (details in Section \ref{utds}).
    \item (\textbf{Test}) One {KK} test set {\textit{$D\textsuperscript{T}\textsubscript{KK}$}}.
    \item (\textbf{Test}) One {UU} test set {\textit{$D\textsuperscript{T}\textsubscript{UU}$}}.
    
\end{itemize}

 Following the evaluation process defined in \citep{Neal1,Oza1}, we split a standard classification dataset (e.g., CIFAR10) into a {KK} part {\textit{$D\textsubscript{KK}$}} and a {UU} part {\textit{$D\textsubscript{UU}$}}. {\textit{$D\textsubscript{UU}$}} is left aside and only its test set counterpart {\textit{$D\textsuperscript{T}\textsubscript{UU}$}} is used for evaluation along with {\textit{$D\textsuperscript{T}\textsubscript{KK}$}}. {\textit{$D\textsuperscript{T}\textsubscript{UU}$}} is used only for testing and our network remains unaware of the classes contained in {\textit{$D\textsuperscript{T}\textsubscript{UU}$}}. A {Known Unknown Trainer} set {\textit{$D\textsubscript{KUT}$}} is mined from another publicly available benchmark dataset \textit{$D\textsubscript{x}$}. {Classes contained in either {\textit{$D\textsubscript{KK}$}} or {\textit{$D\textsuperscript{T}\textsubscript{UU}$}} are removed from \textit{$D\textsubscript{x}$} so that {\textit{$D\textsubscript{KUT}$}} does not contain any overlapping classes ($D\textsubscript{KK} \cap \textit{D\textsuperscript{T}\textsubscript{UU}} \cap D\textsubscript{KUT} = \varnothing $).} Figure \ref{fig:dsets} provides an overview of the dataset splits.


\begin{table}
\caption{An empirical analysis of OSRNet's performance on different {\textit{$D\textsubscript{KUT}$}} selection criteria or threshold $\textit{T}$. It is evident that images from \textit{$D\textsubscript{x}$} with $P\geq 80\%$ perform best as {\textit{$D\textsubscript{KUT}$}}. The average entropy is also the lowest around this point. 
}

	\begin{center}
	\resizebox{.48\textwidth}{!}{%
		\begin{tabular}{c|cc|ccccc}
\hline
\multirow{ 3}{*}{{\textbf{KK}}}
 &&\multicolumn{6}{c}{\textbf{\textit{$D\textsubscript{x}$}} (AUROC \%)}\\
\cline{2-8}

 & Probability  & Entropy & CIFAR100 & Indoor67 & Caltech256 & T- & F-MNIST 
\\

& Threshold (\%)&   $H(X)$  &&&& ImageNet & $+$ ADBase
\\
\cline{1-8}

\multirow{ 5}{*}{\textbf{CIFAR10}} 
&60& 1.65 & 75.60 & 69.25 & 78.65 & 78.23 &-\\
&70& 1.41 & 78.85 & 75.02 & 84.30 & 82.30 &-\\
&  \cellcolor{gray!25} {80} & \cellcolor{gray!25} 1.10 & \cellcolor{gray!25} 89.25 & \cellcolor{gray!25} 83.82 & \cellcolor{gray!25} 93.15 & \cellcolor{gray!25} 88.40 & \cellcolor{gray!25} -\\
&90& 1.21 & 86.32 & 81.40 & 89.66 & 85.49 &-\\

\hdashline
\multirow{ 5}{*}{\textbf{CIFAR+10}}
&60 & 1.71 &-& 72.69 & 80.20 & 80.60 &-\\
&70 & 1.45 &-& 79.05 & 88.54 & 84.25 &-\\
&  \cellcolor{gray!25}{80} & \cellcolor{gray!25} 0.98 & \cellcolor{gray!25} - & \cellcolor{gray!25} 85.76 & \cellcolor{gray!25} 95.69 & \cellcolor{gray!25} 91.40 & \cellcolor{gray!25} -\\
&90 & 1.16 &-& 82.85 & 92.36 & 88.71 &-\\

\hdashline

\multirow{ 5}{*}{\textbf{CIFAR+50}}
&60& 1.55 &-& 71.15 & 83.69 & 77.06 &-\\
&70 & 1.51 &-& 79.30 & 90.04& 84.55 &-\\
& \cellcolor{gray!25} {80} & \cellcolor{gray!25} 1.18 & \cellcolor{gray!25} - & \cellcolor{gray!25} 83.60 & \cellcolor{gray!25} 94.60 & \cellcolor{gray!25} 89.78 & \cellcolor{gray!25} -\\
&90 & 1.24 &-& 80.95 &92.69 & 87.45 &-\\

\hdashline

\multirow{ 5}{*}{\textbf{T-ImageNet}} 
&60& 1.92 & 60.88 & 55.01 & 62.22 &- &-\\
&70& 1.64 & 69.35 & 65.25 &  73.41&- &-\\
& \cellcolor{gray!25} {80} & \cellcolor{gray!25} 1.36 & \cellcolor{gray!25} 74.55 & \cellcolor{gray!25} 70.95 & \cellcolor{gray!25} 77.10 & \cellcolor{gray!25} - &\cellcolor{gray!25} -\\
&90& 1.48 & 72.98 & 68.07 &  75.94 &- &-\\
\hdashline

\multirow{ 5}{*}{\textbf{MNIST}} 
&60 & 1.33 &-&-&-&-&93.42\\
&70 & 1.18 & - & - &  - &- & 95.36\\
& \cellcolor{gray!25} {80} & \cellcolor{gray!25} 0.91 & \cellcolor{gray!25} - & \cellcolor{gray!25} - & \cellcolor{gray!25} - & \cellcolor{gray!25} - & \cellcolor{gray!25} 98.96\\
& 90 & 1.08 & - & - &  - &- & 96.25\\

\hdashline

\multirow{ 5}{*}{\textbf{SVHN}} 
&60 & 1.47 &-&-&-&-&88.40\\
&70 & 1.21 & - & - &  - &- & 91.13\\
& \cellcolor{gray!25} {80} & \cellcolor{gray!25} 1.02  & \cellcolor{gray!25} - & \cellcolor{gray!25} - & \cellcolor{gray!25} - & \cellcolor{gray!25} - & \cellcolor{gray!25} 92.66\\
&90 & 1.14 & - & - &  - &- & 90.69\\

\hline

\end{tabular}
}
\end{center}

\label{table:t5}
\end{table}

OSRNet consists of two parts: a base CNN is responsible for classifying an instance ($x$) regardless of its distribution, i.e., {KK} or {UU}. On the other hand, the CS augmented to the base CNN outputs a score indicating whether $ x \in {\textit{D\textsubscript{KK}}}$ or $ x \notin {\textit{D\textsubscript{KK}}}$ (see Figure \ref{fig:os2}). In other words, even if an input belongs to a class that our network has not seen during training, the CNN will output a label. It is up to the CS to decide whether to accept or reject it. To put these formally, an input image $x$ triggers the $N$-way CNN in OSRNet ($N$ is the number of classes) to produce a probability $y\textsubscript{i}$ ($i \in $ \{$1...N$\}) for each of the classes. The CS outputs a score $S$  for each $x$, indicating the confidence of the input being a {UU} instance. The final OSRNet output $OP$ can be expressed using Equation \ref{eq:e1}.

\begin{equation}\label{eq:e1}
OP=\left\{\begin{array}{ll}{\max P\left(y_{i} | x\right)} & {\text { if } S <\delta},\\  x \notin {\textit{D\textsubscript{KK}}} & {\text { else }}\end{array}\right.
\end{equation}
\quad \quad \quad \quad \quad \quad \quad \quad where, $i \in $ \{$1...N$\}

As shown later, an optimally chosen cut-off value $\delta$ can be applied on $S$ to reject {UU} instances.

As discussed in Section \ref{sec:introduction} and depicted in Figure \ref{fig:heatBar}, the probability scores of a CNN are confined within the visually similar classes when the input belongs to {\textit{$D\textsubscript{KK}$}}. However, the probability distribution does not follow the same trend for unknown instances. Therefore, the intermediate layer features of a CNN should exhibit differences for the {KK} and unknown classes. Since we aim to exploit this behavioural distinction, the choice of the feature layer is important. 
In contrast to a single Fully Connected Layer (\textit{FC\textsubscript{Soft}}) used in traditional ResNet (we call it ResNet\textsubscript{1FC}) \citep{resnet1}, we add one additional FC Layer (\textit{FC\textsubscript{1}}) prior to \textit{\textit{FC\textsubscript{Soft}}} and train this ResNet (we call it ResNet\textsubscript{2FC}) on {\textit{$D\textsubscript{KK}$}} from scratch. A detailed analysis on the impact of different ResNet variants and OSRNet's performance is provided in Section \ref{perform}.

\begin{figure}
\begin{center}
   \includegraphics[width =1 \linewidth]{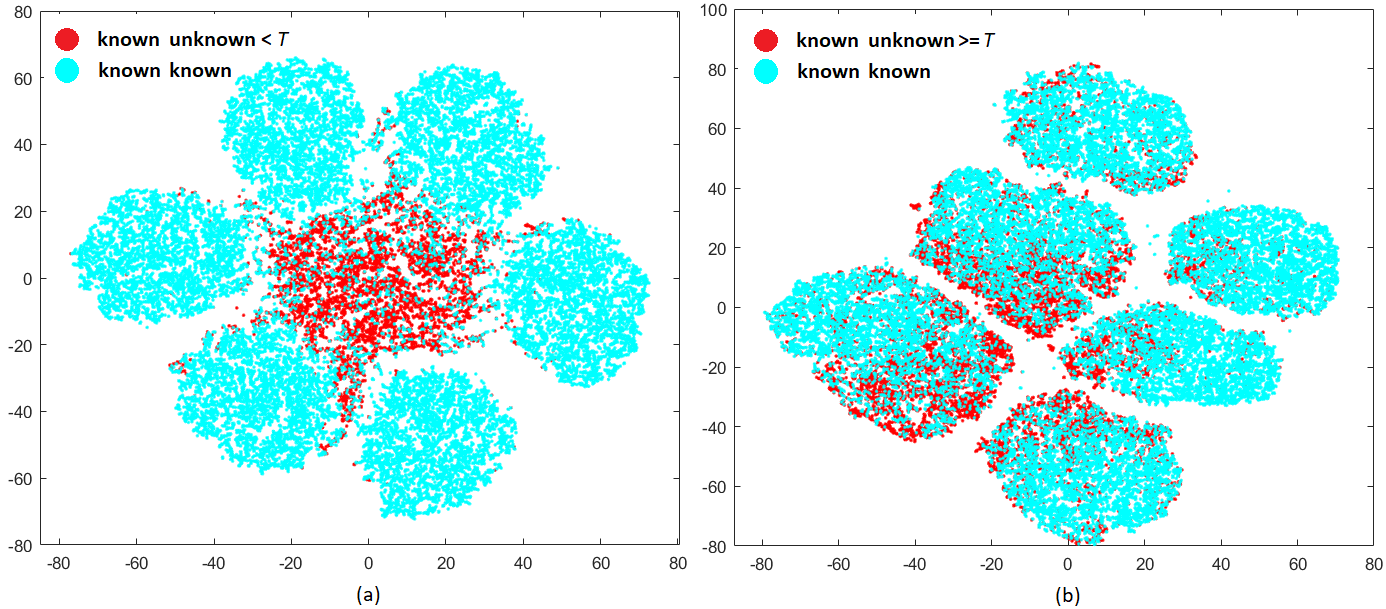}
\end{center}
   \caption{t-SNE plots of \textbf{(a)} \textit{\textit{FC\textsubscript{1}}} features from all images in {\textit{$D\textsubscript{KK}$}} (cyan) and those in \textit{$D\textsubscript{x}$} (red) inducing probability $P <$ some threshold $T$ and \textbf{(b)} $P\geq T$. It is visually evident that separating {KKs} from {KUs} in (b) is more difficult. OSRNet trained to seperate {KKs} from {KUs} in (b) can identify relatively easier {KUs} in (a) as well. An effective probability threshold \textit{T} is chosen based on finding the maximum point on a cubic polynomial curve (see Figure \ref{fig:curve}).}
\label{fig:tsne1}
\end{figure}

\begin{figure}
\begin{center}
   \includegraphics[width=.9\linewidth]{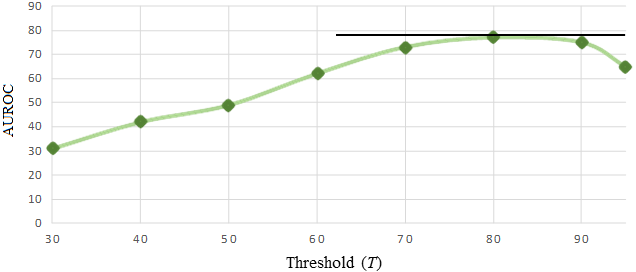}
\end{center}
   \caption{AUROC vs $T$ curve based on sample points for Tiny ImageNet as {\textit{$D\textsubscript{KK}$}} and Caltech256 as \textit{$D\textsubscript{x}$}. A probability threshold $T$ is calculated from the maximum point on the curve. This threshold $T$ calculation is repeated for each {\textit{$D\textsubscript{KK}$}} reported in this work.}
\label{fig:curve}
\end{figure}


\subsection{{Known Unknown Trainer Dataset Mining}}\label{utds}
{
After splitting a standard classification dataset into {\textit{$D\textsubscript{KK}$}} and \textit{$D\textsubscript{UU}$}, ResNet\textsubscript{2FC} is trained on {{\textit{$D\textsubscript{KK}$}}} and remains completely unaware of the classes in \textit{{$D\textsubscript{UU}$}}. 
To mine \textit{{\textit{$D\textsubscript{KUT}$}}}, another publicly available benchmark dataset \textit{$D\textsubscript{x}$} is fed to a ResNet\textsubscript{2FC} trained on {\textit{$D\textsubscript{KK}$}}. From all the images in \textit{$D\textsubscript{x}$}, only the ones inducing probability greater than a threshold are mined for \textit{{\textit{$D\textsubscript{KUT}$}}}, i.e., $\forall{x}$ $\exists{(x)}$ $(P(x) > T)$ (see Figure \ref{fig:tsne1}). Choosing an effective \textit{T} is vital for OSRNet's performance. To select such a $T$, it is important to analyze how different $T$ values fare against OSRNet's performance. Table \ref{table:t5} has a compilation of increasing $T$ against Area Under ROC curve or $AUROC$ score of OSRNet. It is evident that for the given sample points, $T \approx 80\%$ is a good choice across datasets. Calculating OSRNet's performance on every possible $T$ is tedious. Therefore, to pick a $T$, we fit a cubic polynomial curve for all the collected sample points across datasets (e.g., $(70, 78.85)$ is a data point for CIFAR10 ({\textit{$D\textsubscript{KK}$}})-CIFAR100 (\textit{$D\textsubscript{KU}$}) combination). }
The curve is represented by Equation \ref{eq:equa1} (A, B, C, and D are coefficients).

\begin{equation}
F(T)=A+BT+C T^{2}+D T^{3}
\label{eq:equa1}
\end{equation}

A degree two quadratic function has a higher error rate than a cubic one and hence we fit a cubic curve approximated by Equation \ref{eq:equa1} (see Figure \ref{fig:curve}). With the help of first and second-order derivatives, the maximum point on the curve is found where the gradient of a tangent to the curve $F(T)$ is $0$. We fit an $AUROC$ vs $T$ curve for all the {\textit{$D\textsubscript{KK}$}}s with Caltech256 as the \textit{$D\textsubscript{x}$} (as will be explained in the next section, Caltech256 is the best \textit{$D\textsubscript{x}$}). For each {\textit{$D\textsubscript{KK}$}} and \textit{$D\textsubscript{x}$}, a threshold $T$ is calculated from the curve. 

Key characteristics of this curve (Figure \ref{fig:curve}) are explained below:
\begin{itemize}
    \item A smaller $T$ allows a large number of images from \textit{$D\textsubscript{x}$} to qualify for {\textit{$D\textsubscript{KUT}$}}. A considerably small $T$ might allow the entire \textit{$D\textsubscript{x}$} set to qualify for {\textit{$D\textsubscript{KUT}$}} (i.e., $D_x \approx {D_{KUT}}$ if $T \rightarrow 0$). As a result, {\textit{$D\textsubscript{KUT}$}} gets populated with sub-optimal imagery along with high probability inducing ones. OSRNet's performance (AUROC), upto a certain point, improves with increasing $T$. 
    \item OSRNet's performance experiences a downward trend followed by the peak ($T \approx 80\%$). This is because too high of a $T$ value leaves very few images from \textit{$D\textsubscript{x}$} to be eligible for {\textit{$D\textsubscript{KUT}$}}. Such a {\textit{$D\textsubscript{KUT}$}} is inadequate as training dataset for OSRNet. Hence, OSRNet's performance drops as the meagre training data in {\textit{$D\textsubscript{KUT}$}} leads to overfitting.
    \item OSRNet's performance peaks around $T \approx 80\%$ for all the datasets. Such a $T$ offers the best trade-off between the number and quality of images mined for {\textit{$D\textsubscript{KUT}$}}.
\end{itemize}

\begin{algorithm}
\caption{: {${D_{KUT}}$} Construction Process\\\textbf{Input:} Images ($x_n$) from $D_x$, $n \in N$, $N$ is the total number of images in $D_x$.
 \\\textbf{Output:} $x_n$ either selected or discarded for {${D_{KUT}}$}  .}
\label{algo1}
\begin{algorithmic}[1]
\STATE {Choose a benchmark dataset as the base \textit{$D_x$}}

\FOR {all $x_n \in D_x$, where $n = \{{1,2,...N\}}$}

\STATE Feed $x_n$ to $ResNet_{2FC}$
\STATE $ p = max(Probability(x_n))$
\IF {$p > \textit{T}$, where \textit{T} is an optimal threshold found from the curve in Equ. \ref{eq:equa1} }
\STATE {${{D_{KUT}}} \leftarrow x_n $}
\ELSE
\STATE  {Discard $x_n$}
\ENDIF
\ENDFOR
\end{algorithmic}
\end{algorithm}

Algorithm \ref{algo1} summarizes the overall {\textit{$D\textsubscript{KUT}$}} mining process, which is used to teach OSRNet the essence of the {UU} world.
For numeric datasets such as MNIST and SVHN as {\textit{$D\textsubscript{KK}$}}, the options for an ideal \textit{$D\textsubscript{x}$} are limited because natural object classes do not work well as \textit{$D\textsubscript{x}$} for such numeric {\textit{$D\textsubscript{KK}$}}s. In this work, Fashion MNIST \citep{fashion} and ADBase \citep{arabic} are used in conjunction as the base \textit{$D\textsubscript{x}$}.

\subsection{Training OSRNet}
 Training OSRNet involves a series of steps. The overall training workflow is depicted in Figure \ref{fig:os2}, and the steps are explained below. 
 \begin{itemize}

\item ResNet\textsubscript{2FC} is used to extract \textit{FC\textsubscript{1}} features from the same {\textit{$D\textsubscript{KK}$}} it was initially trained on. All these features are labeled as $0$. 

\item \textit{{\textit{$D\textsubscript{KUT}$}}} images (mined following the process in Section \ref{utds}) are fed to the same ResNet\textsubscript{2FC} and \textit{FC\textsubscript{1}} features are extracted. These features are labeled as $1$. 
\item The CS (an ANN with one output node) is trained with these two sets of \textit{FC\textsubscript{1}} features (from {{\textit{$D\textsubscript{KK}$}}} and \textit{{\textit{$D\textsubscript{KUT}$}}}) to teach the difference between the {KK} and the {UU}. As the CS is capable of distinguishing relatively difficult borderline {\textit{$D\textsubscript{KUT}$}} features from {\textit{$D\textsubscript{KK}$}} features, it can identify rest of the relatively easier {KUs} in \textit{$D\textsubscript{x}$}.
\item {Finally,} this trained CS is augmented to the corresponding (\textit{FC\textsubscript{1}}) layer of the pre-trained ResNet\textsubscript{2FC}. OSRNet functions as a single unit at inference time.

\end{itemize}
Algorithm \ref{algo2} summarizes the OSRNet training process.
As the CS takes care of {UU} detection, ResNet\textsubscript{2FC} within OSRNet maintains the original classification accuracy on the test fold of {\textit{$D\textsubscript{KK}$}}. At test time, OSRNet not only produces a class label but also provides a confidence score suggesting how likely an input is to be {UU}.

\begin{algorithm}
\caption{: OSRNet Training\\\textbf{Input:} Individual images from ${D_{KUT}}$ (referred $x_i$) and ${D_{KK}}$ (referred $x_j$).
 \\\textbf{Output:} Trained $OSRNet$  .}
\label{algo2}
\begin{algorithmic}[1]
\FOR {all $x_i \in {D_{KUT}}$, where $i = \{{1,2,....N\}}$}

\STATE Feed $x_i$ to $ResNet_{2FC}$
\STATE ${Feature_{KUT}[i]} \leftarrow FC_1$
\STATE ${Feature_{KUT\_{ label}}[i]} \leftarrow 0$
\ENDFOR

\FOR {all $x_j \in {D_{KK}}$, where $j = \{{1,2,....N\}}$}

\STATE Feed $x_j$ to $ResNet_{2FC}$
\STATE ${Feature_{KK}[j]} \leftarrow FC_1$
\STATE ${Feature_{KK\_ label}[j]} \leftarrow 1$
\ENDFOR

\STATE  {Train Confidence Subnetwork ($CS$) with ${\{Feature_{KUT},Feature_{KUT\_ label}\}}$ and ${\{Feature_{KK},Feature_{KK\_ label}\}}$}
\STATE {Augment $CS$ to $ResNet_{2FC}$}
\end{algorithmic}
\end{algorithm}

\subsection{Why OSRNet works?}

\begin{figure}
\begin{center}
   \includegraphics[width=.92\linewidth]{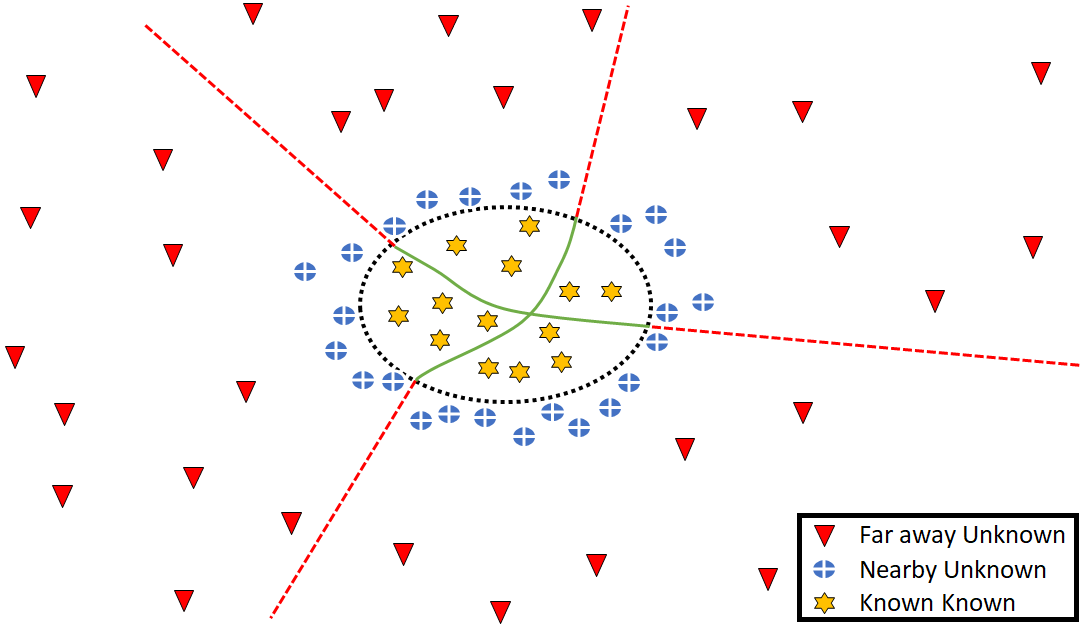}
\end{center}
   \caption{A conceptual illustration of the {\textit{$D\textsubscript{KUT}$}} image characteristics. {\textit{$D\textsubscript{KUT}$}} consists of only the nearby unknowns and OSRNet is trained to detect these difficult to identify unknowns. This way, the far-away unknowns are automatically detected without being used as part of the training. For OSRNet, less training data is required and better accuracy is achieved.}
\label{fig:explain}
\end{figure}

 Since perceiving high dimensional space is difficult for humans, conceptual illustrations are widely used for better understanding \citep{dube1,tanay}. Here, with the help of Figure \ref{fig:explain}, we explain why the proposed {\textit{$D\textsubscript{KUT}$}} and OSRNet work well in OSR. 
 
 The asterisks within the dotted oval resemble {\textit{$D\textsubscript{KK}$}}, and classification within this oval is quite accurate. The negative space or the region beyond the encapsulating oval accommodates all the unknown images (as mentioned earlier, we refer to {KUs} $+$ {UUs} as unknowns), i.e., nearby unknowns (circle), and far away unknowns (triangle). Traditional CNN's performance lack robustness in the negative space since the decision boundaries (dotted red) extrapolate to infinity \citep{Goodfellow1} without any precision. In this work, we only use high probability ($P > T$) inducing images in {\textit{$D\textsubscript{KUT}$}} to represent the {UU}. Such images (blue circles) in {\textit{$D\textsubscript{KUT}$}} reside close to the {KK} distribution (the dotted oval). We argue that a deep classifier capable of treating the borderline {\textit{$D\textsubscript{KUT}$}} instances (blue circles) as unknown can easily identify relatively easier far-away unknowns (triangles). It is understandably a challenging task to collect all the borderline unknowns perfectly encapsulating the {KK} distribution. 

 In this work, we strive to maximize the participation of such borderline images in {\textit{$D\textsubscript{KUT}$}}. A couple of follow-up questions still remain: 
 
\begin{itemize}
\item (\textbf{Q1}) For a certain {\textit{$D\textsubscript{KK}$}}, which \textit{$D\textsubscript{x}$} should we choose to mine {\textit{$D\textsubscript{KUT}$}}?
\item (\textbf{Q2}) Can a {\textit{$D\textsubscript{KUT}$}} with fewer data outperform some \textit{$D\textsubscript{x}$} it was mined from (although ${\textit{$D\textsubscript{KUT}$}} \subset \textit{$D\textsubscript{x}$}$)? 
\end{itemize}
\textbf{Answer to Q1. } We have experimented with four public datasets: CIFAR100, Indoor67, Clatech256, and Tiny ImageNet as \textit{$D\textsubscript{x}$}. It is evident from Table \ref{table:t5} that Caltech256 performs best as \textit{$D\textsubscript{x}$} across datasets and Indoor67 comes out last. One intriguing aspect is comparing the visual characteristics of the best performing \textit{$D\textsubscript{x}$} against others. Is it visually similar to {\textit{D\textsubscript{KK}}} ($\textit{D\textsubscript{x}} \approx {\textit{D\textsubscript{KK}}}$) or is it quite different (we show quantitative distance between different data distributions in Section \ref{discussion})? For example, bus is a visually similar class to truck compared to some indoor images. It turns out, Indoor67 is not a good candidate for \textit{D\textsubscript{x}} as it consists of only interior images (e.g., office room, bedroom, and auditorium) with little to no visual similarity to the ${\textit{D\textsubscript{KK}}}$s. On the other hand, Caltech256 performs better as it consists of classes ranging from a variety of animals to different vehicles having greater visual similarity to the {\textit{D\textsubscript{KK}}}s. 

Our suggestion is to use a \textit{$D\textsubscript{x}$} that has classes similar to {\textit{$D\textsubscript{KK}$}}. E.g., if a certain {\textit{$D\textsubscript{KK}$}} consists of different dog breeds, a dataset containing different cat breeds could be a better option as \textit{$D\textsubscript{x}$} rather than using some indoor images like Indoor67 as the \textit{$D\textsubscript{x}$}. We argue that training to separate an apparently more difficult \textit{$D\textsubscript{x}$}, like Caltech256, from {\textit{$D\textsubscript{KK}$}} enables OSRNet to automatically distinguish relatively easier instances at test time.\\ 
\textbf{Answer to Q2. } Experimental analysis shows that the type of images in {\textit{$D\textsubscript{KUT}$}} is more important than the number of images. It is intriguing that despite being a subset of \textit{$D\textsubscript{x}$}, {\textit{$D\textsubscript{KUT}$}} teaches OSRNet the essence of the unknown world better than a much larger in size \textit{$D\textsubscript{x}$}. A detailed explanation of this observation is provided in Section \ref{discussion}.

 To further understand the conceptual explanation and why mining only high probability inducing {KUs} for {\textit{$D\textsubscript{KUT}$}} makes sense, we resort to entropy comparison. In multi-class classification tasks such as ours, a trained CNN provides a class probability for each of the output nodes. The probability distribution ($D_P$) generated by the CNN portrays how certain the network is about the classification. When a test instance is classified confidently, one output node exhibits higher ($\approx 1$) probability than the rest. The uncertainty factor is low here as is the entropy. Conversely, when the classifier is susceptible to the input and does not provide a confident probability distribution (e.g., a Uniform Distribution), the uncertainty is high and hence the entropy as well. This suggests that for an input, lower the $D_P$ uncertainty, closer the input is to {\textit{$D\textsubscript{KK}$}}. This uncertainty can be quantified from $D_P$ entropy using Equation \ref{eq:equ2}.
 
 \begin{equation}
H(X)=H\left(p_{1}, \ldots, p_{n}\right)=-\sum_{i=1}^{n} p_{i} \log _{2} p_{i}
\label{eq:equ2}
\end{equation}
 
 In a number of cases, when a trained deep CNN faces {UU} instances, the entropy or $H(X)$ stays low. This generally means those images are adjacent to the {KK} distribution despite actually being {UU}. On the other hand, the rest of the images induce high $H(X)$ which implies such images are generally not adjacent to {\textit{$D\textsubscript{KK}$}} and rightly so. As can be seen from Table \ref{table:t5}, the lowest entropy co-occurs with the best performing $T$. This supports the usage of only high probability inducing images ($P > T$) in {\textit{$D\textsubscript{KUT}$}} since such a {\textit{$D\textsubscript{KUT}$}} would be closer to {\textit{$D\textsubscript{KK}$}} and harder to distinguish resulting in a tight but efficient decision boundary.

\section{Experiments}\label{experiments}
\subsection{Datasets and splits}

\textbf{MNIST, SVHN, CIFAR10. } MNIST \citep{lecun2010mnist} is a digits dataset containing 60,000 training and 10,000 test images. Each digit from 0 to 9 denotes a class. All the images are grayscale and have a resolution of $28 \times 28$. SVHN \citep{svhn} is also a digits dataset, but the images are collected from Google Street View cameras capturing house numbers (from 0 to 9 as well). This dataset is considered harder than MNIST and contains $32 \times 32$ colour images. It has 73,257 training and 26,032 test instances. CIFAR10 \citep{cifar1} also has 10 classes with $32 \times 32$ colour images. It has 50,000 training and 10,000 test images of different objects (e.g., cat and dog). 
To train OSRNet individually, each of these three datasets is split into {{\textit{$D\textsubscript{KK}$}}} with six classes and \textit{{$D\textsubscript{UU}$}} with four classes. The split for corresponding test set is same as the training set, i.e., {\textit{$D\textsuperscript{T}\textsubscript{KK}$}} and {\textit{$D\textsuperscript{T}\textsubscript{UU}$}} contain same classes as their training split. For example,  while training on MNIST, six randomly selected classes are used as \textit{{$D\textsubscript{KK}$}} and the other four classes are considered as \textit{{$D\textsubscript{UU}$}}. For each dataset, the `openness' is estimated by Equation \ref{eq:e2} \citep{open1,Neal1}. Greater openness value implies higher difference in the number of classes between {\textit{$D\textsubscript{UU}$}}/{\textit{$D\textsuperscript{T}\textsubscript{UU}$}} and \textit{{$D\textsubscript{KK}$}}/{\textit{$D\textsuperscript{T}\textsubscript{KK}$}}, i.e., the {UU} to {KK} class ratio is higher. It is worth reinstating that no images from {\textit{$D\textsubscript{UU}$}} is used in training OSRNet. Only the test fold {\textit{$D\textsuperscript{T}\textsubscript{UU}$}} is used for testing.

\begin{equation}\label{eq:e2}
\text { openness }=1-\sqrt{\frac{ | \text { N } |}{ | \text { Q}|}}
\end{equation}
Here, $|N|$ and $|Q|$ denote the number of \textit{{$D\textsubscript{KK}$}} classes and the number of total test classes ({\textit{$D\textsuperscript{T}\textsubscript{UU}$}} $+$ {\textit{$D\textsuperscript{T}\textsubscript{KK}$}}) respectively. 
The term $openness$ is measured in percentage where higher value represents greater $openness$.
According to Equation \ref{eq:e2}, $openness$ of CIFAR10, MNIST and SVHN is $22.50\%$.\\
\textbf{CIFAR+10. } Four classes from CIFAR10 are selected as \textit{{$D\textsubscript{KK}$}} and 10 non-overlapping classes from CIFAR100 are selected as \textit{{$D\textsubscript{UU}$}}. $openness$ is $46.50\%$.\\
\textbf{CIFAR+50. } Four classes from CIFAR10 are selected as \textit{{$D\textsubscript{KK}$}} and 50 non-overlapping classes from CIFAR100 are selected as \textit{{$D\textsubscript{UU}$}}. $openness$ is $72.78\%$.\\
\textbf{Tiny ImageNet. } Tiny ImageNet is a subset of ImageNet \citep{imagenet1} comprising 200 classes with $64 \times 64$ colour images. Each of the classes has 500 training and 50 test images. We randomly select 20 classes as \textit{{$D\textsubscript{KK}$}} and 180 classes as \textit{{$D\textsubscript{UU}$}}. $openness$ is $68.38\%$
The class selection process is random, and for each benchmark dataset, we experiment with five iterations and the average is reported (Section \ref{perform}). 

\subsection{{Known Unknown Trainer} Datasets}
We use publicly available benchmark datasets (\textit{$D\textsubscript{x}$}) to mine {\textit{$D\textsubscript{KUT}$}}.\\
\textbf{Caltech256. } As explained earlier, Caltech256 performs best as \textit{$D\textsubscript{x}$} for all non-digit \textit{{$D\textsubscript{KK}$}}s. It has 256 object categories containing 30,607 images. All the overlapping classes among \textit{$D\textsubscript{x}$}, \textit{{$D\textsubscript{KK}$}}, \textit{{$D\textsubscript{UU}$}} are removed to achieve a fair (no prior knowledge about {\textit{$D\textsuperscript{T}\textsubscript{UU}$}}) and sane (no {{\textit{$D\textsubscript{KK}$}}} class is used later in training as {UU}) {\textit{$D\textsubscript{KUT}$}}. \\
\textbf{Fashion MNIST, ADBase. } Fashion MNIST \citep{fashion} has exactly the same attributes as MNIST but contains 10 classes of fashion products. ADBase \citep{arabic} is the Arabic version of MNIST. Images in both of these datasets are of resolution $28 \times 28$ and grayscale. These two datasets are used collectively as the \textit{$D\textsubscript{x}$} for both of the numeric datasets, i.e., when \textit{{$D\textsubscript{KK}$}} belongs to either MNIST or SVHN.

\subsection{Training CNN}\label{implement}

\textbf{CIFAR10, CIFAR+10, CIFAR+50, SVHN. } 
Each of these datasets has $32 \times 32$ colour images. After splitting the datasets into {\textit{$D\textsubscript{KK}$}} and {\textit{$D\textsubscript{UU}$}} following the process described in the previous section, a CNN classifier is trained on \textit{{$D\textsubscript{KK}$}}. ResNet\textsubscript{2FC} (with depth 20 \citep{resnet1}) is used as the CNN in OSRNet. The first $3 \times 3$ convolution layer is followed by six $3 \times 3 \times 16$, six $3 \times 3 \times 32$ and six $3 \times 3 \times 64$ convolution layers where the first two dimensions represent the filter size and the third dimension stands for the number of filters. The SoftMax output is preceded by an $N$-way ($N$ is the number of classes in {\textit{$D\textsubscript{KK}$}}) \textit{FC\textsubscript{Soft}} layer. \textit{FC\textsubscript{Soft}} layer in turn is preceded by the additional \textit{FC\textsubscript{1}} layer.\\
\textbf{Tiny ImageNet. } ResNet\textsubscript{2FC} (with depth 32 \citep{resnet1}) is used to train on \textit{{$D\textsubscript{KK}$}} of Tiny ImageNet. It has six $3 \times 3 \times 16$, $3 \times 3 \times 32$, six $3 \times 3 \times 64$, six $3 \times 3 \times 128$, and $3 \times 3 \times 256$ convolution layers. The \textit{FC\textsubscript{Soft}} layer here as well, is preceded by an additional \textit{FC\textsubscript{1}} layer.
Since images in this dataset has a higher spatial resolution, a deeper variant of ResNet\textsubscript{2FC} is used. 

The greater distinctiveness of the additional \textit{FC\textsubscript{1}} features from ResNet\textsubscript{2FC} compared to solitaire \textit{FC\textsubscript{Soft}} features in ResNet aids OSRNet's open set recognition performance. The benefits come at a negligible increase in the total number of network parameters ($\approx 3.5 \%$). Detailed results are provided in Section \ref{perform}.

\textbf{MNIST. } For MNIST, we train a plain CNN on {\textit{$D\textsubscript{KK}$}} consisting of three $3 \times 3$ convolution blocks with respective filter numbers of {8, 16, and 32}. These layers are followed by two FC layers (128 unit \textit{FC\textsubscript{1}} and $N$-way \textit{FC\textsubscript{Soft}}) and the SoftMax output layer. ResNet architecture is not adopted for MNIST as a plain CNN network works just fine with competitive classification accuracy ($99.66\%$). 

 We adopted stochastic gradient descent (SGD), data shuffling before every epoch while training and data augmentation (horizontal flip and translation). Minibatch size of 128 is used for all the datasets except MNIST (8,192). Adaptive dropout \citep{hossain1} is followed to avoid overfitting. A multi-class cross-entropy loss ($\mathcal{L}_m$) is used as the objective function (Equation \ref{e7}).

\begin{equation}\label{e7}
\operatorname{\mathcal{L}_m}=-\sum_{i=1}^{N} \sum_{j=1}^{K} \mathrm{t}_{i j} \ln y_{i j}
\end{equation}
where $N$ is the number of samples, $K$ is the number of classes, $t_{ij}$ denotes that the $i\textsuperscript{th}$ sample belongs to the $j\textsuperscript{th}$ class, and $y_{ij}$ is the output for sample $i$ for class $j$, which effectively is the value from the SoftMax function, i.e., it is the probability that the network associates the $i\textsuperscript{th}$ input with class $j$.

\begin{table*}
\caption{AUROC performance (\%) comparison of different {OSR and OOD methods.}}

\label{table:t1}
	\begin{center}
	\resizebox{.92\textwidth}{!}{%
		\begin{tabular}{ccccccc}
\hline

\textbf{Method} & \textbf{CIFAR10} & \textbf{CIFAR+10} & \textbf{CIFAR+50} & \textbf{MNIST} & \textbf{SVHN} & \textbf{Tiny ImageNet} \\
\hline
\rowcolor{gray!12} SoftMax  & $67.70$ & $81.60$ & $80.50$ & $97.80$ & $88.60$ & $57.70$\\

\rowcolor{gray!12} OpenMax (Bendale et al. \citep{Bendale1}) & $69.50$ & $81.70$ & $79.60$ & $98.10$ & $89.40$ & $57.60$\\

\rowcolor{gray!12} G-OpenMax (Ge et al. \citep{Ge1}) & $67.50$ & $82.70$ & $81.90$ & $98.40$ & $89.60$ & $58.00$ \\

\rowcolor{gray!12} OSRCI (Neal et al. \citep{Neal1}) & $69.90$ & $83.80$ & $82.70$ & $98.80$ & $91.00$ & $58.60$ \\

\rowcolor{gray!12} C2AE (Oza et al. \citep{Oza1}) & $89.50$ & $95.50$ & $93.70$ & $98.90$ & $92.20$ & $74.80$ \\

 \rowcolor{red!50} OE (\cite{HendrycksDeepAnomaly1})  & $63.50$ & $73.40$ & $71.20$ & $93.11$ & $81.40$ & $44.90$\\

 \rowcolor{red!50} ODIN (\cite{odin})  & $64.10$ & $75.10$ & $72.20$ & $96.10$ & $81.30$ & $47.10$\\

 \rowcolor{red!50} G-ODIN (\cite{odin2}) & $68.80$ & $79.10$ & $73.20$ & $97.40$ & $84.70$ & $47.30$ \\

 \rowcolor{gray!30} Proposed Method (\textit{$D\textsubscript{x}$}) & ${90.40}\pm{0.08}$ & ${94.88}\pm{0.12}$ & ${93.50}\pm{0.11}$ & ${98.35}\pm{0.09}$ & ${91.22}\pm{0.14}$ & ${75.70}\pm{0..20}$ \\

\rowcolor{gray!30} Proposed Method (\textit{{\textit{$D\textsubscript{KUT}$}}}-ResNet1FC) & ${91.66}\pm{0.05}$ & ${94.95}\pm{0.08}$ & ${94.03}\pm{0.10}$ & ${98.44}\pm{0.05}$ & ${91.45}\pm{0.10}$ & ${76.80}\pm{0.16}$ \\

 \rowcolor{gray!30} \textbf{Proposed Method (\textit{{\textit{$D\textsubscript{KUT}$}}}-ResNet2FC)} & $\textbf{93.15}\pm{0.04}$ & $\textbf{95.69}\pm{0.06}$ & $\textbf{94.60}\pm{0.07}$ & $\textbf{98.96}\pm{0.04}$ & $\textbf{92.66}\pm{0.09}$ & $\textbf{77.10}\pm{0.13}$ \\

\hline
\end{tabular}
}
\end{center}
\end{table*}

\begin{table}
\centering
\caption{Classification accuracy (\%) comparison of contemporary OSR methods. OSRNet\textsubscript{1} is the variant of OSRNet with ResNet\textsubscript{1FC}.}
\begin{adjustbox}{max width=.4\textwidth}
\begin{tabular}{ccccc}

\multicolumn{5}{c}{\textbf{CIFAR10} \textit{{$D\textsubscript{KK}$}}} 

\\
\cline{1-5}
 OSRNet\textsubscript{1} & OSRNet & OpenMax & G-OpenMax & OSRCI\\

80.20 & 82.91 & 80.10 & 81.60 & 82.10\\
\cline{1-5}
\multicolumn{5}{c}{\textbf{MNIST} \textit{{$D\textsubscript{KK}$}}}
\\

99.50 & 99.72 & 99.50 & 99.60 & 99.60\\
\cline{1-5}
\multicolumn{5}{c}{\textbf{SVHN} \textit{{$D\textsubscript{KK}$}}}
\\

 94.61 & 95.95 & 94.70 & 94.80 & 95.10\\
\cline{1-5}
\\
\end{tabular}
\end{adjustbox}

\label{table:t2}
\end{table}


\subsection{Training CS}

As depicted earlier in Figure \ref{fig:os2}, CS is an ANN subnetwork within OSRNet and is responsible for detecting {UUs} while ResNet\textsubscript{2FC} classifies the instance.
Features (for \textit{{$D\textsubscript{KK}$}} and \textit{{\textit{$D\textsubscript{KUT}$}}}) extracted from ResNet\textsubscript{2FC}'s \textit{FC\textsubscript{1}} layer are fed to CS as the training data. Binary class labels are supplied as the ground truth and variable learning rate gradient descent (GDX) with momentum is used as the training method. We train CS with two hidden layers and the number of hidden units ($H$) per layer is between the number of inputs and outputs \citep{ann1}. For 128 dimensional \textit{FC\textsubscript{1}} features, $H$ is set to {64, 128,  and 256} while for 256 dimensional \textit{FC\textsubscript{1}} features, $H$ is set to {128, 256, and 512}. A binary cross-entropy loss function ($\mathcal{L}_b$) is used following Equation \ref{eq:bce}.

\begin{equation}\label{eq:bce}
\text { $\mathcal{L}_b$ }=-\frac{1}{N} \sum_{i=1}^{M} T_{i} \log \left(Y_{i}\right)
\end{equation}
where $M$ is the total number of responses in $Y$, $N$ is the total number of observations in $Y$, $Y_i$ is the network output, and $T_i$ is the target value. 

For each configuration, 10-fold training is conducted. Later, the best average score yielding CS is augmented to the corresponding ResNet\textsubscript{2FC} to form OSRNet. It is observed that, CS with $H$ as 64 performs best for 20-depth ResNet\textsubscript{2FC} and CS with $H$ as 128 performs best for 32-depth ResNet\textsubscript{2FC}. Hence the reported results in Table \ref{table:t1} are derived using these two ensembles.

\subsection{Optimum Confidence Cut-Off $\delta$ Estimation}\label{optimum}

Area Under the ROC curve provides an overview of a binary classifier's performance. A greater AUROC magnitude indicates better performance. However, when the classifier is deployed in a real-life application, it is expected to rule out negative samples based on a cut-off value $\delta$. A classifier's ultimate accuracy largely depends on the choice of $\delta$. An optimal $\delta$ ensures maximum correct classification with minimum error. Such a $\delta$ can be estimated from the optimal operative point on the ROC curve. There are multiple ways of finding such a point on ROC curve that will provide a good compromise between FPR (1 – specificity) and TPR (sensitivity). However, in this work, we follow the optimal slope intersection method \citep{roc3_swets,roc2_zweig} where an initial slope $S\textsubscript{op}$ is calculated in the ROC curve from the misclassification cost using Equation \ref{eq:cost}.
 
 \begin{equation}\label{eq:cost}
S\textsubscript{op}=\frac{\operatorname{C}(P | N)-\operatorname{C}(N | N)}{\operatorname{C}(N | P)-\operatorname{C}(P | P)} \times \frac{N}{P}
\end{equation}
 $C(N|P)$ is the cost of misclassifying a positive class as a negative class. $C(P|N)$ is the cost of misclassifying a negative class as a positive class. In our case, the cost of any form of misclassification is equally penalized ($C(N|P) = C(P|N) = 0.5$). On the other hand, there is no penalty for a correct classification ($C(N|N) = C(P|P) = 0$).
 
$P = True Positive + False Negative$ 

$N = True Negative + False Positive$.

 
A line with the slope $S\textsubscript{op}$ is dragged from the upper-left corner of the ROC plot (where $FPR = 0, TPR = 1$) down and to the right. The first intersection point of the slope line and the ROC curve is the optimal operating point. The value which ensures the optimal cut-off point is selected as the $\delta$ when OSRNet is deployed.

\begin{figure}
\begin{center}
   \includegraphics[width=1\linewidth]{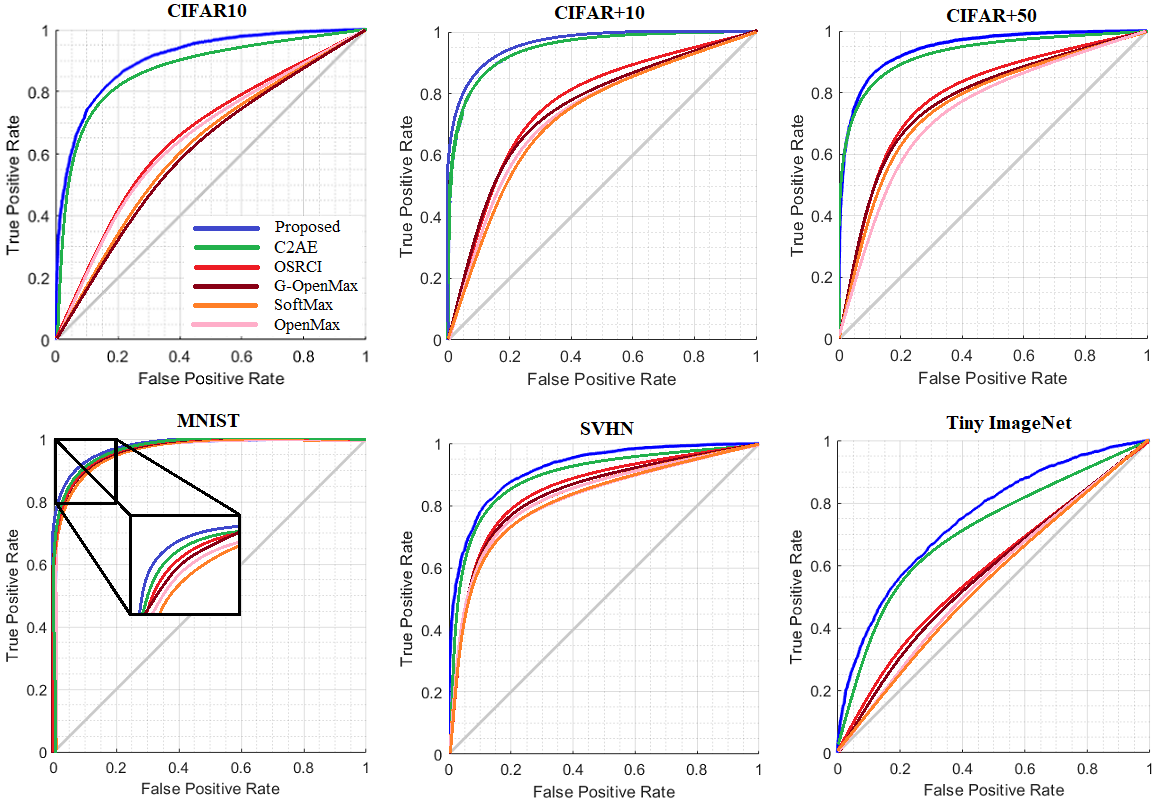}
\end{center}
   \caption{AUROC performance comparison between proposed OSRNet and other methods in the literature. OSRNet outperforms all the existing methods on all six benchmark datasets. }
\label{fig:roc}

\end{figure}

\begin{table}
\caption{Classification accuracy (\%) comparison among traditional ResNet\textsubscript{1FC}, ResNet\textsubscript{2FC}, and ResNet\textsubscript{3FC}. ResNet\textsubscript{2FC} performs better on the benchmark classification task ({without KK-UU split)}. Tiny ImageNet does not have labeled test set hence not shown here. }
	\begin{center}
	\resizebox{1\columnwidth}{!}{%
		\begin{tabular}{c|c|c|c}
\hline
&\multicolumn{3}{c}{\textbf{CIFAR10}} \\
\cline{2-4}
\multirow{7}{*}{\textbf{Classification Accuracy (\%)}} & ResNet\textsubscript{1FC} & ResNet\textsubscript{2FC} & ResNet\textsubscript{3FC}\\
& 90.24 & 90.27 & 88.95\\
&\multicolumn{3}{c}{\textbf{MNIST}}\\
\cline{2-4}
& PlainCNN\textsubscript{1FC} & PlainCNN\textsubscript{2FC} & ResNet\textsubscript{3FC}\\
& 98.87 & 99.66 & 98.66\\
&\multicolumn{3}{c}{\textbf{SVHN}}\\
\cline{2-4}
& ResNet\textsubscript{1FC} & ResNet\textsubscript{2FC} & ResNet\textsubscript{3FC}\\
& 94.77 & 95.29 & 93.05

\\
\hline
\end{tabular}
}
\end{center}

\label{table:t4}
\end{table}

 
\begin{table*}[!h]
\centering
\caption{AUROC performance (\%) comparison of different benchmark datasets as the base \textit{$D\textsubscript{x}$} for choosing {\textit{$D\textsubscript{KUT}$}}. The number of images inside \textit{$D\textsubscript{x}$} is listed in the column header and the number of images ultimately qualifying for {\textit{$D\textsubscript{KUT}$}} is provided inside the parenthesis in each row. Interestingly, {\textit{$D\textsubscript{KUT}$}} - although a subset of \textit{$D\textsubscript{x}$} and much smaller in size, works consistently better than using entire \textit{$D\textsubscript{x}$} as {\textit{$D\textsubscript{KUT}$}} across all four datasets.}

	\begin{center}
	\resizebox{.7\textwidth}{!}{%
		\begin{tabular}{c|c|c|c|c|c|c|c|c}
\hline
\multirow{ 3}{*}{\textbf{{KK}}}
 &\multicolumn{8}{c}{\textbf{{UU}}}\\
\cline{2-9}
& \multicolumn{2}{c|}{CIFAR100}
& \multicolumn{2}{c|}{Indoor67}
& \multicolumn{2}{c|}{Caltech256}
& \multicolumn{2}{c}{Tiny ImageNet}
\\
\cline{2-9}
 & \textit{$D\textsubscript{x}$}  & {\textit{$D\textsubscript{KUT}$}} ($>$ \textit{T}) & \textit{$D\textsubscript{x}$} & {\textit{$D\textsubscript{KUT}$}} ($>$ \textit{T}) & \textit{$D\textsubscript{x}$}  & {\textit{$D\textsubscript{KUT}$}} ($>$ \textit{T}) & \textit{$D\textsubscript{x}$}  & {\textit{$D\textsubscript{KUT}$}} ($>$ \textit{T}) 
\\
& (all 46k) & & (all 16k) & & (all 28k) & & (all 100k) &
\\
\hline
CIFAR10 & 86.10 & 89.25 (25k) & 82.03 & 83.82 (9k) & 90.40 & 93.15 (18k) & 87.54 & 88.40 (64k)
\\
CIFAR+10 & -&-& 83.65 & 85.76 (9.5k) & 94.88 & 95.69 (19k) & 88.45 & 91.40 (62k)
\\
CIFAR+50 & - & - & 82.33 & 83.60 (9k) & 93.50 & 94.60 (18k) & 86.94 & 89.78 (60k)\\
Tiny ImageNet & 72.06 & 74.55 (23k) & 69.50 & 70.95 (7.5k) & 75.70 & 77.10 (16k) & - & -
\\
\hline
\end{tabular}
}

\label{table:t3}
\end{center}
\end{table*}

 \section{Performance Comparison}\label{perform}

 An open set recognizer performs two tasks simultaneously: it has to identify an input either as {KK} or {UU} and classify it correctly if {KK}. While accuracy is a good metric to gauge classification performance, {UU} detection performance should be evaluated on a metric that takes True Positive Rate (TPR) and False Positive Rate (FPR) into account. Therefore, we provide evaluation results separately with AUROC as the detection metric and TOP-1 accuracy as the classification metric.
 OSRNet outperforms other works in the literature under both metrics- AUROC (see Table \ref{table:t1} and Figure \ref{fig:roc}) and classification accuracy (see Table \ref{table:t2}). 
 
 {
 Class Conditioned Auto-Encoder (C2AE) \citep{Oza1} comes close to OSRNet in terms of AUROC performance on a number of datasets. C2AE follows a 3-stage training scheme. First, a dataset is randomly divided into \textit{D\textsubscript{KK}} and \textit{D\textsubscript{UU}}. An encoder (a CNN minus the SoftMax output) is trained on \textit{D\textsubscript{KK}}. A decoder is augmented to the encoder output and is trained to reconstruct any given input. The entire network (when the encoder-decoder ensemble works as one unit) is designed to reconstruct any \textit{D\textsubscript{KK}} instance as precisely as possible. However, it is designed to poorly reconstruct \textit{D\textsubscript{UU}} instances so that the reconstruction error is high for {UUs} at inference time. Finally, at inference time, reconstruction error is calculated for each input. \textit{D\textsubscript{KK}} inputs are expected to be correctly reconstructed. Therefore, the reconstruction error for \textit{D\textsubscript{KK}} instances should be close to zero. For \textit{D\textsubscript{UU}} instances, however, the reconstruction will have a significant error compared to its \textit{D\textsubscript{KK}} counterpart. This way, the magnitude of the error is exploited to determine whether an instance belongs to the {UU} or not. 
 As advocated in \cite{HendrycksDeepAnomaly1}, we attribute OSRNet's better task performance to the use of an effective, diverse, and real Known Unknown Trainer dataset over encoder-decoder based generative models.
 }

As mentioned earlier, detecting {UUs} is one side of OSR. Maintaining accuracy on classifying {\textit{D\textsubscript{KK}}} images is the other side of it.
Compared to the traditional ResNet\textsubscript{1FC}, our modified ResNet with two FC layers (ResNet\textsubscript{2FC}) performs better on OSR task (See Table \ref{table:t1}). ResNet\textsubscript{2FC} not only benefits our OSR performance, it also ameliorates the classification accuracy (see Table \ref{table:t2}). ResNet\textsubscript{2FC} performs better than ResNet\textsubscript{1FC} on the standard classification task (see Table \ref{table:t4}). We argue that the increased depth of the FC layers in ResNet\textsubscript{2FC} provides richer features and boosts classification accuracy as well. However, using more than two FC layers do not work as well as using two (see our empirical analysis in Table \ref{table:t4}). 

{\textbf{Comparison with OOD methods.}
Outlier Exposure (OE) proposed in \cite{HendrycksDeepAnomaly1} emphasizes on training with real and diverse KU datasets. To represent outliers  (when $D\textsubscript{KK} \subset $\textbf{ CIFAR10}), the work \cite{odin} trains on all the images in the `80 million tiny image' dataset \cite{torralba80} while we only use 18k images. In this case, our novelty lies in not naively using entire datasets as outliers. As for ODIN \cite{odin} and Generelized ODIN \cite{odin2}, both use temperature scaling while the latter not requiring any additional outlier training data. However, as shown in Table \ref{table:t1}, OSRNet performs better than the above mentioned OOD methods under the OSR evaluation protocol. We attribute this to the different loss landscapes for these two paradigms which although similar, are not exactly same \cite{Engstrom1}.
}

 \section{Discussion}\label{discussion}
 To perform the OSR task, OSRNet is trained with a specially mined dataset {\textit{$D\textsubscript{KUT}$}}. We discussed in Section \ref{utds} the mining process of {\textit{$D\textsubscript{KUT}$}} and the type of images that work well as \textit{$D\textsubscript{x}$}. However, the significance of the size of {\textit{$D\textsubscript{KUT}$}} (or the number of images) is also worth discussing.
 Our experimental results show that the type of images is more important than the number of images. Our proposed method leaves only around half of the images (16k) from Caltech256 (\textit{$D\textsubscript{x}$}) into {\textit{$D\textsubscript{KUT}$}}. However, it performs better than using the entire Caltech256 (28k) as the {\textit{$D\textsubscript{KUT}$}} (Table \ref{table:t3}). To understand the underlying reason behind this, we need to go back to the conceptual illustration provided in Figure \ref{fig:explain}. In order to detect {UUs} at test time, the best way is to collect a set of borderline {KUs} and train the classifier to draw a decision boundary between this and {\textit{$D\textsubscript{KK}$}}. Anything beyond this decision line should be treated as unknown. Using an entire \textit{$D\textsubscript{x}$} set as {\textit{$D\textsubscript{KUT}$}} (\cite{saviour1}) is not a good idea since such a {\textit{$D\textsubscript{KUT}$}} would contain both the borderline (circles) and far-away (triangles) images. We argue that most of the {KU} instances from \textit{$D\textsubscript{x}$} that do not make it to the {\textit{$D\textsubscript{KUT}$}} can be automatically detected at test time if our classifier draws a line in between {\textit{$D\textsubscript{KUT}$}} (borderline ones) and {\textit{$D\textsubscript{KK}$}}. 
 The following section elaborates on the experimental analysis.\\

\textbf{CIFAR100. } If we use all 46k images from CIFAR100 as the {\textit{$D\textsubscript{KUT}$}}, the AUROC scores come out as 86.10\% and 72.06\% for CIFAR10 and Tiny ImageNet respectively. Whereas, if the images inducing probability greater than 80\% are used in {\textit{$D\textsubscript{KUT}$}}, the score becomes 89.25\% (25k) and 74.55\% (23k) for CIFAR10 and Tiny ImageNet respectively. \\
\textbf{Indoor67. } If we use all 16k images from Indoor67 as the {\textit{$D\textsubscript{KUT}$}}, the AUROC scores come out as 82.03\%, 83.65\%, 82.33\%, and 69.50\% for CIFAR10, CIFAR+10, CIFAR+50, and Tiny ImageNet respectively. Whereas, if the images inducing probability greater than 80\% are used as {\textit{$D\textsubscript{KUT}$}}, the score becomes 83.82\% (9k), 85.76\% (9.5k), 83.60\% (9k), and 70.95\% (7.5k) for CIFAR10, CIFAR+10, CIFAR+50, and Tiny ImageNet respectively.\\
\textbf{Caltech256. } If we use all 28k images from Caltech256 as the {\textit{$D\textsubscript{KUT}$}}, the AUROC scores come out as 90.40\%, 94.88\%, 93.50\%, and 75.70\% for CIFAR10, CIFAR+10, CIFAR+50, and Tiny ImageNet respectively. Whereas, if the images inducing probability greater than 80\% are used as the {\textit{$D\textsubscript{KUT}$}}, the score becomes 93.15\% (18k), 95.69\% (19k), 94.60\% (18k), and 77.10\% (16k) for CIFAR10, CIFAR+10, CIFAR+50, and Tiny ImageNet respectively.\\
\textbf{Tiny ImageNet. } If we use all 100k images from Tiny ImageNet as the {\textit{$D\textsubscript{KUT}$}}, the AUROC scores come out as 87.54\%, 88.45\%, and 86.94\% for CIFAR10, CIFAR+10 and CIFAR+50 respectively. Whereas, if the images inducing probability greater than 80\% are used as the {\textit{$D\textsubscript{KUT}$}}, the score becomes 88.40\% (64k), 91.40\% (62k), and 89.78\% (60k) for CIFAR10, CIFAR+10, and CIFAR+50 respectively.

Table \ref{table:t3} does not include digit-based MNIST and SVHN datasets in the {KK} column as such other datasets are scarcer. Non-digit datasets with natural images (e.g., Caltech256) do not work well at all as the \textit{$D\textsubscript{x}$} for MNIST and SVHN.
It can be summarized that Caltech256 performs best as \textit{$D\textsubscript{x}$} for mining {\textit{$D\textsubscript{KUT}$}} because of its similarity with {\textit{$D\textsubscript{KK}$}} along with class diversity.
\\
{
\textbf{Train and Test Data Distributions. } }
{
 It is possible that even after explicitly removing all the overlapping classes, there could still be an implicit overlap between \textit{$D^{T}\textsubscript{UU}$} and \textit{$D\textsubscript{KUT}$} because of the nature of these datasets. However, the core of our proposed method revolves around the fact that \textit{$D\textsubscript{KUT}$} lies close to the {\textit{$D\textsubscript{KK}$}} distribution. Therefore, our network's decision boundary around the {\textit{$D\textsubscript{KK}$}} distribution is fairly tight. Hence, OSRNet is capable of successfully identifying {UUs} regardless of the distribution of \textit{$D^{T}\textsubscript{UU}$}. A conceptual illustration is provided in Figure \ref{fig:explain}.
}
{
To show OSRNet performs consistently with or without any implicit overlap between \textit{$D\textsubscript{KUT}$} and \textit{$D^{T}\textsubscript{UU}$}, we expand our test datasets for further experimentations. In addition to the default \textit{$D^{T}\textsubscript{UU}$}, we test on two separate \textit{$D^{T}\textsubscript{UU}$}s according to the following setup: }
 
\begin{itemize}
\item {\textit{$D\textsubscript{KK} \subset CIFAR10$} (default train split).} 
    \item {\textit{$D^{T}\textsubscript{UU}\_S = $} SVHN testset.}

    \item {\textit{$D^{T}\textsubscript{UU}\_I = $} Indoor67 testset.}

    \item {\textit{$D^{T}\textsubscript{UU}\_C \subset CIFAR10 $} testset (default unknown test split). }
    \item {\textit{$D^{T}\textsubscript{UU}\_C \cap D\textsubscript{KK} = \emptyset $}. }

\end{itemize}

{
CIFAR10 contains 10 different classes such as bird, dog and six randomly selected classes are used as the \textit{$D\textsubscript{KK}$}. Indoor67 dataset contains 67 interior classes, e.g., bedroom, office, and garage, while SVHN contains 10 classes belonging to digits (0-9). These two datasets do not visually or semantically overlap with the \textit{$D^{T}\textsubscript{UU}$}\_C. In fact, samples from both \textit{$D^{T}\textsubscript{UU}$}\_S and \textit{$D^{T}\textsubscript{UU}$}\_I are easier to identify as outliers compared to \textit{$D^{T}\textsubscript{UU}$}\_C owing to their greater distance from \textit{$D\textsubscript{KK}$}. Table \ref{table:dist} supports our claim, i.e., OSRNet performs better on distant and non-overlapping unknowns. On \textit{$D^{T}\textsubscript{UU}$}\_S and \textit{$D^{T}\textsubscript{UU}$}\_I, OSRNet produces 5.15\% and 1.80\% greater AUROC score respectively compared to \textit{$D^{T}\textsubscript{UU}$}\_C. Table \ref{table:mmd} reinforces our argument regarding the inter distribution distance measured using Maximum Mean Discrepancy (MMD) \cite{mmd1}.
}

\begin{table}
\centering
\caption{{\textit{$D^{T}\textsubscript{UU}$} from the left out set of CIFAR-10 is the most challenging testset compared to \textit{$D^{T}\textsubscript{UU}$}s drawn from non-overlapping SVHN and Indoor67 testsets.}}
\begin{adjustbox}{max width=.45\textwidth}

\begin{tabular}{ccccc}
& \multicolumn{3}{c}{{D\textsubscript{KK}} $\subset $\textbf{ CIFAR10} } 
\\
\cline{1-4}
\\
 &&AUROC (\%) &\\
\cline{2-4}
\\
 & {\textit{$D^{T}\textsubscript{UU}$}\_C}  & {\textit{$D^{T}\textsubscript{UU}$}\_S}  & {\textit{$D^{T}\textsubscript{UU}$}\_I}  \\
\cline{2-4}
\\
\rowcolor{red!50} C2AE (\citep{Oza1}) & $89.50$ & $92.88$ & $88.59$\\

\rowcolor{red!50}  OSRNet & ${93.15}$ & ${98.30}$ & ${94.95}$ \\
\cline{1-4}
\\
\end{tabular}

\end{adjustbox}

\label{table:dist}
\end{table}

{
 MMD can approximate the distance between the underlying distribution of two image datasets based on the Reproducing Kernel Hilbert Space (RKHS) \cite{mmd2,mmd3}. We used the final FC layer features from OSRNet Backbone to represent the dataset distributions.
Let $X=\left\{x_{1}, \ldots, x_{n_{1}}\right\}$ and $Y=\left\{y_{1}, \ldots, y_{n_{2}}\right\}$ be two datasets with distributions $\mathcal{P}$ and $\mathcal{Q}$. The empirical distance between $\mathcal{P}$ and $\mathcal{Q}$, according to MMD, is
$$
\operatorname{Dist}(\mathrm{X}, \mathrm{Y})=\left\|\frac{1}{n_{1}} \sum_{i=1}^{n_{1}} \phi\left(x_{i}\right)-\frac{1}{n_{2}} \sum_{i=1}^{n_{2}} \phi\left(y_{i}\right)\right\|_{\mathcal{H}}
$$
where $\mathcal{H}$ is a universal RKHS \cite{steinwart1}, and $\phi$ :
$\mathcal{X} \rightarrow \mathcal{H}$
}

\begin{table}
\centering
\caption{{Both \textit{$D^{T}\textsubscript{UU}$}\_S and \textit{$D^{T}\textsubscript{UU}$}\_I have greater distance from {\textit{$D\textsubscript{KK}$}} compared to the default testset \textit{$D^{T}\textsubscript{UU}$}\_C. This complements our findings in Table \ref{table:dist} where OSRNet performs better on distant testsets.  }}
\begin{adjustbox}{max width=.47\textwidth}

\begin{tabular}{ccccc}

& \multicolumn{4}{c}{{D\textsubscript{KK}} $\subset $\textbf{ CIFAR10} } 
\\
\cline{1-5}
\\
& {\textit{$D\textsubscript{KK}$}} $\rightarrow$  {\textit{$D\textsubscript{KUT}$}} &
{\textit{$D\textsubscript{KK}$}} $\rightarrow$  {\textit{$D^{T}\textsubscript{UU}$}\_C} &{\textit{$D\textsubscript{KK}$}} $\rightarrow$ {\textit{$D^{T}\textsubscript{UU}$}\_S} & {\textit{$D\textsubscript{KK}$}}  $\rightarrow$  {\textit{$D^{T}\textsubscript{UU}$}\_I} \\

&   & CIFAR10 & SVHN & Indoor67\\

\cline{2-5}
\\
\rowcolor{red!50}MMD & 0.18 & 0.23 & 0.59 & 0.44 \\
\cline{1-5}
\\
\end{tabular}

\end{adjustbox}

\label{table:mmd}
\end{table}

{
It is evident from Table \ref{table:mmd} that {\textit{$D\textsubscript{KUT}$}} indeed resides close to {{\textit{$D\textsubscript{KK}$}}}. It is also apparent that visually and semantically distant datasets are indeed far away from {{\textit{$D\textsubscript{KK}$}}} and hence, easier to detect. This is analogous to our findings in Table \ref{table:dist} where OSRNet performs better on Both \textit{$D^{T}\textsubscript{UU}$}\_S and \textit{$D^{T}\textsubscript{UU}$}\_I than \textit{$D^{T}\textsubscript{UU}$}\_C.
Our experimental results also complement the conceptual visualisation in Figure \ref{fig:explain}.
}

 \section{Conclusion}\label{conclusion}

Open set recognition is one of the pressing issues deep image classifiers are faced with. In this work, we analyzed CNNs' behaviour on both {KK} and {UU} datasets and proposed a network that is adept at the OSR task. Instead of using synthetic images, we have demonstrated that a trainer dataset mined from publicly available datasets can represent the unknown better. OSRNet functions as a single end-to-end unit at inference time. It outperforms contemporary OSR techniques on a number of benchmark datasets. 
We have also discussed the reasons behind OSRNet's success in the given task through conceptual decision boundaries and comparative entropy analysis. {There is still room for improvement in terms of generalisation of our training approach. To be more specific, we used heuristical assumptions to select a source dataset to mine the Known Unknown Trainers. We believe a distance based metric, such as Maximum Mean Discrepancy, can be used to justify the selection of a certain source dataset. This can be investigated further in future. Besides, images detected as unknowns by OSRNet can be further cognized. This will open up avenues for unsupervised image labelling as well.}

\bibliographystyle{model2-names}
\bibliography{refs}

\end{document}